**Spiral Complete Coverage Path Planning Based on Conformal Slit Mapping in Multi-connected Domains**

Changqing Shen[1], Sihao Mao[1], Bingzhou Xu[1], Ziwei Wang[1], Xiaojian Zhang[1*], Sijie Yan[1] and Han Ding[1]

[1] State Key Laboratory of Intelligent Manufacturing Equipment and Technology, School of Mechanical Science and Engineering, Huazhong University of Science and Technology, Wuhan, 430074, China

**Corresponding author:**

Xiaojian Zhang, State Key Laboratory of Intelligent Manufacturing Equipment and Technology, School of Mechanical Science and Engineering, Huazhong University of Science and Technology, No. 1307, Luoyu Street, Hongshan District, Wuhan, 430074, China

E-mail: xjzhang@hust.edu.cn

**Abstract**

The generation of smoother and shorter spiral complete coverage paths in multi-connected domains is a crucial research topic in path planning for robotic cavity machining and other related fields. Traditional methods for spiral path planning in multi-connected domains typically incorporate a subregion division procedure that leads to excessive subregion bridging, requiring longer, more sharply turning, and unevenly spaced spirals to achieve complete coverage. To address this issue, this paper proposes a novel spiral complete coverage path planning method using conformal slit mapping. It takes advantage of the fact that conformal slit mapping can transform multi-connected domains into regular disks or annuluses without the need for subregion division. Firstly, a slit mapping calculation technique is proposed for segmented cubic spline boundaries with corners. Secondly, a spiral path spacing control method is developed based on the maximum inscribed circle radius between adjacent conformal slit mapping iso-parameters. Thirdly, the spiral coverage path is derived by offsetting iso-parameters. Numerical experiments indicate that our method shares a comparable order of magnitude in computation time with the traditional PDE-based spiral complete coverage path method, but it excels in optimizing total path length, smoothness, and spacing consistency. Finally, we performed experiments on cavity milling and dry runs to compare the new method with the traditional PDE-based method in terms of machining duration and steering impact, respectively. The comparison reveals that, with both algorithms achieving complete coverage, the new algorithm reduces machining time and steering impact by 12.34% and 22.78%, respectively, compared with the traditional PDE-based method.

connected Domain, Cavity Machining

# 1. Introduction

*1.1 Problem Description*

In the fields of cavity machining (Lei et al., 2023; Makhanov, 2022), 3D printing (Zhai & Chen, 2019), unmanned vehicles (Wu et al., 2019), the Complete Coverage Path Planning (CCPP) algorithm is widely used. After fully perceiving the domain to be covered, the CCPP problem can be defined as determining one curve or several bridged curves of a specified width capable of covering an entire known simply connected or multi-connected regions (Khan et al., 2017). The chosen curve width is determined by the operational capacities of the actuators, such as the radius of the milling cutter (Szaroleta et al., 2023).

While actuators capable of omnidirectional motion can navigate along trajectories with sharp turns to avoid excessive uncovered regions (Kapadia et al., 2011), they still face constraints such as impact resistance and the differential steering ability of the actuators. Deceleration is necessary for the actuators to negotiate non-smooth corners (Yang et al., 2002). Additionally, in the field of 3D printing, non-continuous printing trajectories can also adversely affect print strength (Cam & Gunpinar, 2023). Therefore, smooth and continuous spiral trajectory coverage patterns have garnered extensive attention in the academic community. Various types of spiral trajectory coverage patterns have found widespread applications in fields such as cavity machining (Bieterman & Sandstrom, 2003), 3D printing (Zhai & Chen, 2019) and unmanned vehicles (Wu et al., 2019). But the current challenges faced by spiral trajectory coverage patterns are shared across different applications when applied to complex multi-connected regions. This involves the necessity to decompose the space to be covered into mutually exclusive and exhaustive subregions. Subregions with numerous and small areas require more bridges and spiral coverage with larger, longer, and unevenly spaced turning curvatures (Spielberger & Held, 2013). Therefore, decomposing too many subregions diminishes the advantages of spiral coverage.

Our objective is to propose a novel spiral coverage pattern capable of minimizing the number of subdivided subregions, resulting in a consistent reduction in total length, severity of turns, number of bridges, and addressing inconsistencies in trajectory spacing for spiral complete coverage paths.

*1.2 Related work*

Existing CCPP algorithms can be mainly classified into three categories: the grid map method, the element decomposition method, and the methods aiming to obtain smooth spiral trajectories. The grid map method involves three primary steps: domain meshing, movement cost function definition on the mesh, and determination of the optimal path (Xing et al., 2023; Kyriakakis et al., 2022; Wan et al., 2018; Arkin et al., 2000; Modares et al., 2017; Yehoshua et al., 2016; Hassan & Liu, 2020). This method has been comprehensively investigated across various domains. For example, Jalil, Modares et al. experimentally demonstrated that reducing the number and degree of steering in the path can decrease the energy consumption of drone flight, They also mentioned that the number of grid cells to obtain the globally optimal energy consumption path through intelligent algorithms is limited by computational capacity (Modares et al., 2017) . The Spiral STC algorithm (Gabriely & Rimon, 2002) creates a cycle-free tree-like structure by utilizing spiral scanning and backtracking mechanisms on a grid. It generates a bridge-free complete coverage trajectory based on this structure. While the algorithm efficiently achieves complete coverage by following predefined motion rules, but the algorithm cannot guarantee fewer turns in the path. Grid-based methods encounter three common challenges. Firstly, the pursuit of a globally optimal path using most grid-based methods is NP-hard (Xing et al., 2023; Kyriakakis et al., 2022; Wan et al., 2018; Arkin et al., 2000; Modares et al., 2017; Yehoshua et al., 2016; Hassan & Liu, 2020). Due to the computational overhead of NP-hard problems, augmenting grid density becomes intricate, making it difficult to achieve precise modeling of environmental boundaries. Secondly, grid-based methods enforce constraints on turning angles, often restricted to multiples of 15° (Wan et al., 2018; Hassan & Liu, 2020), resulting in paths that lack smoothness. Lastly, in scenarios with complex boundary conditions, the paths generated by this method often necessitate frequent turning or bridging, leading to extended trajectories for the actuator and the potential for performing high-cost motions.

The element decomposition method also involves three steps. The first step is to divide a complex multi-connected region into several simple subregions. Then, space-filling curves are employed to fill these subregions, followed by considering the bridging methods between subregions. The related work encompasses the Morse decomposition method (Acar et al., 2002), which divides the region into trapezoids by solving the critical points of boundaries and sweep lines. In scenarios with folded boundaries, it is challenging to design suitable sweep lines to reduce the number of cell decompositions. The polygon subdivision method (Makhe & Frank, 2010) and the geodesic power Voronoi tessellation method (Bhattacharya et al., 2013) during the cell decomposition, even small-volume holes will generate new subregion boundaries, thereby increasing the number of subregions. Similarly, Yu-Yao Lin et al. decomposes 3D surfaces based on critical trajectories of holomorphic quadratic differentials, for each additional hole on the curved surface, the number of subregions decomposed by this method will increase by three. (Lin et al., 2017), and other methods (Oksanen & Visala, 2009). Thus, when the multi-connected region is complex with ruffle-shaped boundaries, it is challenging to find a method that can reasonably reorganize the region into a simply connected subregions in suitable number, leading to the inclusion of multiple bridges in the generated trajectory. Repetitive lines are the most commonly used curves for filling the subregions, but this type of path is less smooth and not necessarily the fastest (Kim & Choi, 2002).

Generating a spiral trajectory is another effective approach for addressing CCPP issues. Wu et al. conducted experiments and observed that, in comparison to the zigzag coverage mode, the spiral coverage mode reduces steering severity, leading to lower energy consumption for the unmanned vehicle's differential steering. This results in a notable 14.5% decrease in total power consumption for the unmanned vehicle to reach complete coverage (Wu et al., 2019). Similarly, Sun et al., in their experiments on surface milling, found that the uninterrupted spiral trajectory reduces the number of tool lifts and sharp transitions in the trajectory. Consequently, when compared with interrupted concentric tool paths and contour-parallel tool paths, there is a reduction in tool lift residuals on the machined surface and residuals at corners (Sun et al., 2016). These experiments underscore the practical significance of the spiral coverage mode in various fields. As a result, in the pursuit of the benefits arising from the smoothness and continuity of trajectories, scholars have proposed numerous methods for generating Spiral Complete Coverage Paths (SCCP). Typical SCCP planning methods include those based on medial axis tree (MAT) (Huang et al., 2020; Abrahamsen, 2019; Spielberger & Held, 2013; Xu et al., 2020; Held & de Lorenzo, 2018), those based on evolutionary iso-parameter curves (Osher & Sethian, 1988; Zou et al., 2014; Zou & Zhao, 2013), those based on partial differential equation (PDE) (Bieterman & Sandstrom, 2003; Oulee et al., 2004; Yang et al., 2003; Chuang & Yang, 2007; Wang et al., 2023; Cam & Gunpinar, 2023; Zhou et al., 2015), and those based on conformal texture mapping (Hu et al., 2015; Sun et al., 2006; Ren et al., 2009; Sun et al., 2016; Liu et al., 2019; Driscoll, 2023). The MAT methods support the generation of spirals on simply connected regions by gradually offsetting along the medial axis. But due to the existence of multiple loops in the medial axis of multi-connected regions, it cannot be directly used for spiral generation in such regions. Therefore, the MAT methods require the prior division of multi-connected regions into several simply connected subregions. For instance, Martin Held et al. divided the multi-connected region into singly connected subregions using the Voronoi diagram and generated spiral paths in each subregion using MAT (Spielberger & Held, 2013; Held & de Lorenzo, 2018). On multi-connected regions, the evolution of the curve from where holes are enclosed by the curve to where holes are not enclosed is a discontinuous process. Therefore, defining a smooth evolving potential field or evolution rules for curves to avoid intersections with holes during the evolution process becomes challenging. Methods based on PDEs and conformal mapping share similar approaches. They can plan spiral trajectories within simple shapes, such as circles (Patel & Lalwani, 2017; Ren et al., 2009; Spielberger & Held, 2013; Sun et al., 2006; Held & de Lorenzo, 2018) or annuli (Wang et al., 2023; Chuang & Yang, 2007; Zhou et al., 2015), and then map these trajectories to spiral paths within complex boundaries. For multi-connected regions, PDE-based methods also have some variations, for example, adding coincident boundaries can convert a multi-connected region into a singly-connected or dually-connected subregion, but this will lead to a longer path and uneven spacing (Abrahamsen, 2019). Jui-Jen Chuang pointed out that a multi-connected region can be divided into several singly-connected and dually-connected subregions, and then the PDE-based method can be used to generate spiral paths in the subregions (Chuang & Yang, 2007; Driscoll, 2023). Some studies on subregion division are based on the research of Jui-Jen Chuang (Wang et al., 2023; Sun et al., 2016), but the path spacing is also uneven when the divided ring width changes dramatically. Besides, since the iso-parameters calculated by the PDE-based method are not smooth and unevenly spaced at the boundary corners, the above method also leads to unsmooth and uneven spacing in the path at the added corners of the subregion boundaries. Moreover, there are some alternative strategies for generating spiral trajectories (Zhuang et al., 2010; Wu et al., 2019). In these methods, trajectories require multiple bridging and exhibit sharp turns in certain regions. This is due to the non-concentric alignment of the iso-parameters of the trajectories when void holes are present in those regions.

Conformal slit mapping (CSM) is a special class of conformal mapping, and has been used in various scenarios, including fluid (Baddoo & Ayton, 2021; Amano et al., 2012), image registration (Sangawi et al., 2021), and signal propagation (Ban et al., 2013). So far, it is not possible to embed a spiral trajectory into a multi-connected region solely through circular or annular conformal mappings (Sun et al., 2006; Liu et al., 2019; Hu et al., 2015; Peng, 2018; Choi, 2021). Considering that CSM can transform multi-connected regions into regular disks or annuluses without subregion division, we believe that CSM can be applied to coverage path planning to solve the problem that the existing method relies on subregion division. Figure 1 shows the contrast between the final path generated in this paper and that produced by traditional methods. Several solution methods to CSM can be found in references (Nasser, 2011; Yunus et al., 2014; Nasser et al., 2013; Nasser, 2015; Wu & Lu, 2023; Crowdy & Marshall, 2006; Ban et al., 2013).

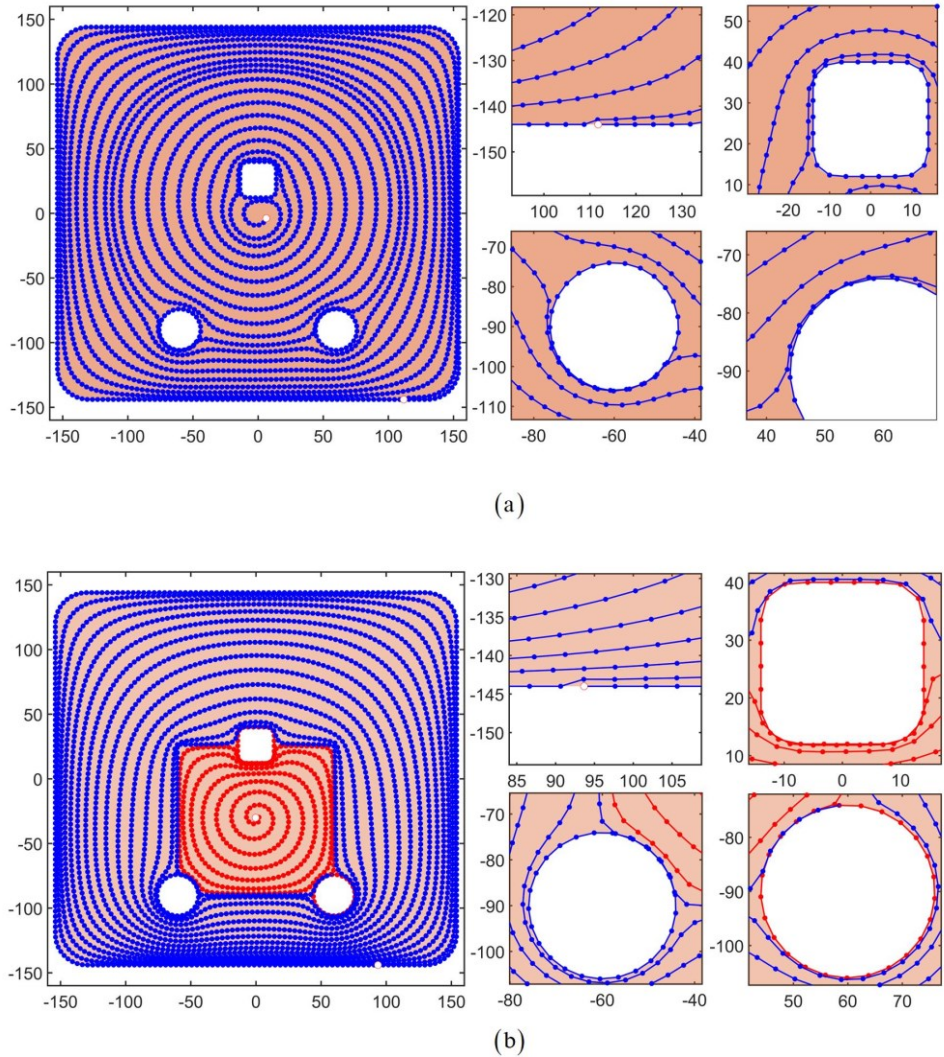

**Figure 1.** Comparison of two types of spiral path generation methods. (a) CSM-based method with path length 12581.84. (b) traditional PDE-based method (Chuang & Yang, 2007; Driscoll, 2023) with path length 14412.83.

*1.3 Primary innovation*

Our primary innovation lies in the pioneering adoption of a CSM-based framework for generating spiral trajectories without bridging. This method eliminates the need to divide sub-regions, produces shorter and fewer corners, and generates spiral complete coverage paths with more uniform spacing. Meanwhile, to ensure complete coverage of the generated path in the application scenario of cavity milling, we propose a CSM iso-parameters spacing control strategy, a CSM iso-parameter spiral offsetting strategy and a strategy for fusing regional boundaries with spiral trajectory.

## 2. Method

Our algorithm mainly consists of three parts, and detailed information will be explained in the corresponding sections:

I. Obtaining a disc or annular slit mapping inverse solver $\omega^{-1}(\cdot)$, which can calculate the corresponding coordinates in the preimage domain based on the coordinates of points within a mapped input disk or annular mapping domain. This corresponds to Section 2.1 and Appendix.

II. Calculating a series of iso-parameters with appropriate spacing based on $\omega^{-1}(\cdot)$. This corresponds to Section 2.2.1.

III. Generating spiral trajectories based on the principle of avoiding the mapping slits, causing iso-parameters to deviate towards adjacent iso-parameters. This corresponds to Section 2.2.2.

*2.1 Two types of conformal slit mapping*

For narrative convenience, this section starts by building symbols that express CSM as $\omega$ and its inverse mapping as $\omega^{-1}$. Let $G$ be a bounded open $m$-multiply ($m > 1$) connected domain with the origin $O \in G$ in the extended complex plane $C \cup \{\infty\}$ and $(m + 1)$-bounds $\partial G = \Gamma = \Gamma_0 \cup \Gamma_1, \dots, \Gamma_{m-1} \cup \Gamma_m$, where $\Gamma_0, \Gamma_1, \dots, \Gamma_m$ are closed Jordan curves. $\Gamma_j\ j = 0,1, \dots, m$ can be smooth or segmentally smooth with $p_j$ corner points. $\Gamma_0$ represents the outer curve enclosing the origin, and all the other inner curves are denoted as $\Gamma_1, \Gamma_2, ..., \Gamma_m$. The domain $G \cup \partial G$ is denoted as $\bar{G}$. For disc slit mapping, we only define the origin $O \in G$, and for annular slit mapping, we define the origin $O \in G$ and a point $Z_1$ inside a specific hole domain. Figure 2 shows the disc slit mapping and annular slit mapping. The former maps the irregular, multi-connected closed domain $\bar{G}$ onto regular unit disc $\bar{D}$ with $\omega(O) = O^*$, and the latter maps $\bar{G}$ onto unit annuli $\bar{A}$ with $\omega(Z_1) = O^*$. Both types of CSM map the boundaries $\Gamma$ of $\bar{G}$ onto circles or circular arc with several radius $R_0, R_1, ..., R_m$, where $R_0 = 1$. For annular slit mapping, we define $\Gamma_1$ as the boundary wraps around the $Z_1$, so, $\Gamma_1$ is mapped to the circular inner boundary of the annular with $R_1 = min\{R_0, R_1, ..., R_m\}$. The position of $O$ in annular slit mapping only introduces a rigid rotation to the mapping result, as shown in Figure 2(e)(f). The CSM is analytic on $G$, indicating that when the points or curves are subjected to the CSM $\omega$ or its inverse $\omega^{-1}$, certain topological properties are preserved, e.g., maintaining the intersection relationships between curves, preserving similarity in point-to-point distances before and after mapping, and retaining the smoothness of curves after mapping. Our path-planning method leverages these properties of the CSM.

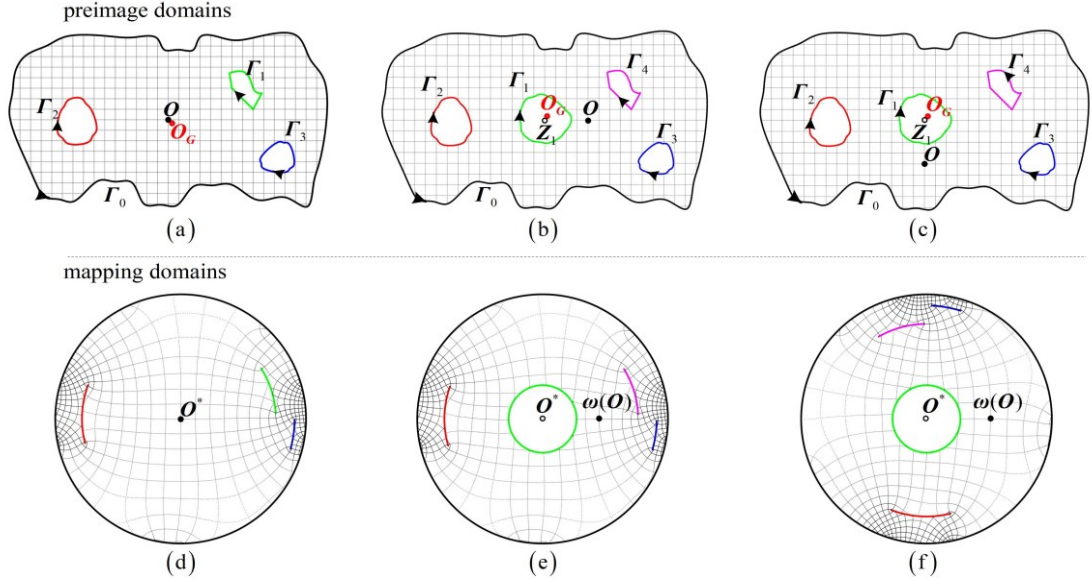

**Figure 2.** Two types of conformal slit mapping. (a) and (d), disc slit mapping preimage and image; (b) and (e), annular slit mapping preimage and image, $O$ is to the right of $Z_1$; (c) and (f), annular slit mapping preimage and image, $O$ is below $Z_1$.

The CSM and its inverse mapping solution method used in this study are consistent with the method in Ref. (Yunus et al., 2014; Nasser et al., 2013; Nasser, 2015; Nasser, 2011) except for the boundary $2\pi$-periodic parameterization part. The solution flow is shown in Figure 3, and more detailed solution procedure is given in Appendix. Considering the specific engineering problems faced by the potential readers of this paper, we introduce segment smooth cubic splines into the boundary $2\pi$-periodic parameterization process, thus, provides a reference for boundary $2\pi$-periodic parameterization in other segment spline forms.

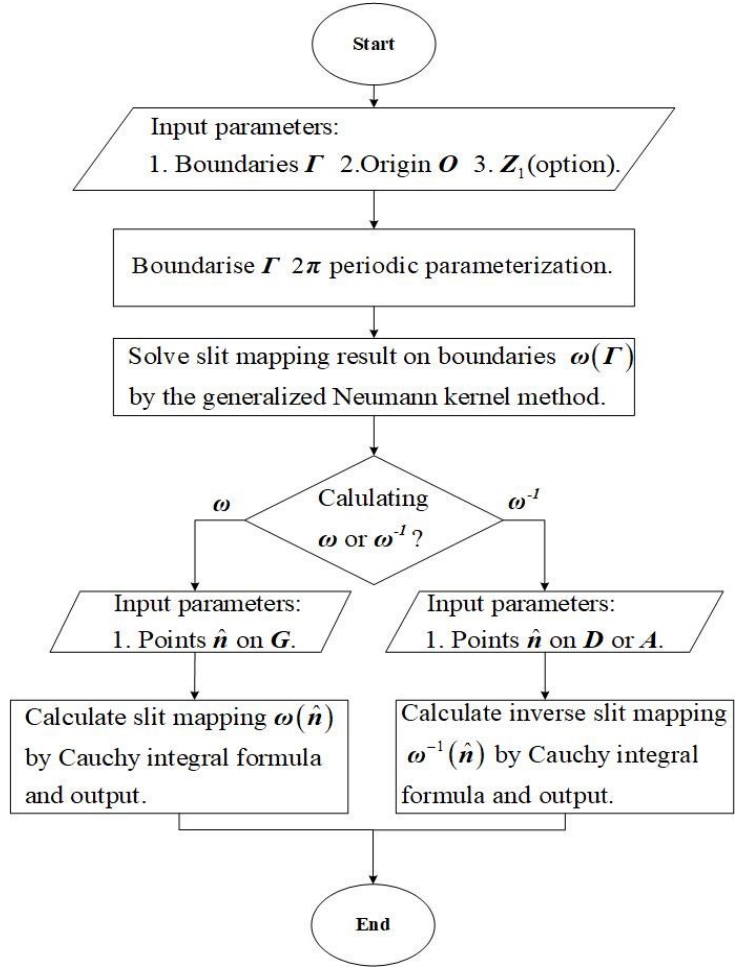

**Figure 3.** Conformal slit mapping and its inverse mapping solution flow.

Assuming domain $\bar{G}$ is a uniformly thick and massed plate, we define the centroid of this plate as $O_G$. The domain to be covered is divided into two cases according to whether $O_G$ is surrounded by a hole. In case I, as shown in Figure 2(a), the domain near the $O_G$ still needs to be covered, so we use disc slit mapping $\omega: \bar{G} \to \bar{D}$ and defines $O$ near the $O_G$. In case II, as shown in Figure 2 (b)(c), the domain near $O_G$ is the hole that does not need to be covered, so we use annular slit mapping $\omega: \bar{G} \to \bar{A}$ and defines $Z_1$ inside the center hole.

*2.2 Spiral path generation*

Traditional spiral trajectory is generated by progressively offsetting a series of concentric iso-parameters toward their adjacent counterparts in a singly-connected or dually-connected domain, and only an upper limit spacing needs to be set between the spiral and itself to ensure complete coverage (Ren et al., 2009; Liu et al., 2019). The two aforementioned cases of CSM can map different multi-connected regions into either a singly-connected disc domain or a dually-connected annular domain with slits. But two problems need to be solved with the combination of offsetting iso-parameters and CSM to generate a spiral trajectory. Firstly, if the spiral trajectory in the mapping domain intersects with the slit, the corresponding spiral trajectory in the preimage domain will pass through the forbidden area. As shown on the left side in Figure 4, there is one intersection point between a spiral and a slit in the mapping domain, and the trajectory of the spiral on the pre-image crosses the obstacle region corresponding to the slit. Secondly, in the preimage domain, the large spacing between the spiral may not necessarily lead to incomplete coverage of the spiral trajectory, because the large spacing may also be caused by the forbidden area sandwiched in the middle of the spiral. As shown on the right side in Figure 4, in the mapping domain, the points on both sides of the slit have a large spacing in the pre-image domain. But this does not imply that the trajectory cannot fully cover, as there are holes in between that do not need to be covered.

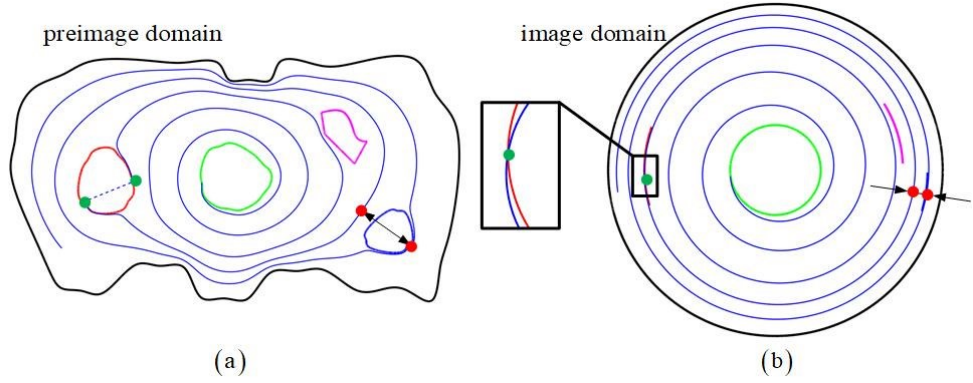

**Figure 4.** Challenges in spiral trajectory generation with offsetting iso-parameters and CSM mapping. (a) The trajectory crossing a hole once in the pre-image domain, along with two points on the trajectory where the spacing is large but does not affect the coverage. (b) The spiral trajectory in the mapping domain, the image of the slit corresponding to the boundary, along with the intersection points of the trajectory with the slit, and two points displaced on either side of the slit along the spiral trajectory.

Therefore, the challenge lies in developing a method to generate a smooth spiral trajectory that avoids crossing holes while maintaining complete coverage. In the following section, our SCCPP method will be introduced.

2.2.1 Iso-parameters spacing control

The concentric circular line clusters on a disk or ring, along with their corresponding line clusters within the pre-image domain, do not intersect each other and cover the entire domain. Therefore, they can be respectively regarded as iso-parameter clusters on the mapping domain and the pre-image domain. When generating a spiral, the offset of the iso-parameters is inconsistent, so there is a difference between the spacing of the spiral trajectory from itself and the spacing between corresponding adjacent iso-parameters. Since such a difference is not obvious, the spacing of iso-parameters can still be taken as a basis for controlling the spacing of spiral trajectories.

Now, a series of iso-parameters will be calculated to control the maximum spacing of the spiral path. As shown in Figure 5, supposing two concentric circles $C_i, C_j$ on mapping space $D$ or $A$ with radius $R_{C_i}, R_{C_j}$, for case I $0 \leq R_{C_i} < R_{C_j} \leq 1$, for case II $R_1 \leq R_{C_i} < R_{C_j} \leq 1$, their preimage curves on $G$ are concentric iso-parameters $L_{C_i} = \omega^{-1}(C_i)$, $L_{C_j} = \omega^{-1}(C_j)$, respectively. We define the domain without an origin and enclosed by $L_C$ as $G_C$ and defines the distance $L_{C_i-C_j}$ between $L_{C_i}$ and $L_{C_j}$ as the radius of the maximum inscribed circle of multi-connected domain $G_{C_i} \cap \complement_{G_{C_j}} \cap G$, where $\complement_{G_{C_j}}$ denotes the complement of $G_{C_j}$. Meanwhile, we define the distance $L_{C-C}$ between $L_C$ and itself as the radius of the maximum inscribed circle of the domain $G_C \cap G$. Based on (Chen & Fu, 2011; Wan et al., 2023), it can be deduced that if we have the milling cutter milling along the domain $G_{C_i} \cap \complement_{G_{C_j}} \cap G$ boundaries and we want all the material on the domain $G_{C_i} \cap \complement_{G_{C_j}} \cap G$ to be removed, the required minimum milling cutter radius is $SR = L_{C_i-C_j}$. As mentioned before, when the spiral is generated by offsetting $L_{C_i}$ and $L_{C_j}$, the minimum cutter radius $C$ that can cover the void between the spiral trajectories is also close to $L_{C_i-C_j}$. Therefore, we set constant $C$ to be equal or slight smaller than the milling cutter radius to control the spacing between adjacent iso-parameters.

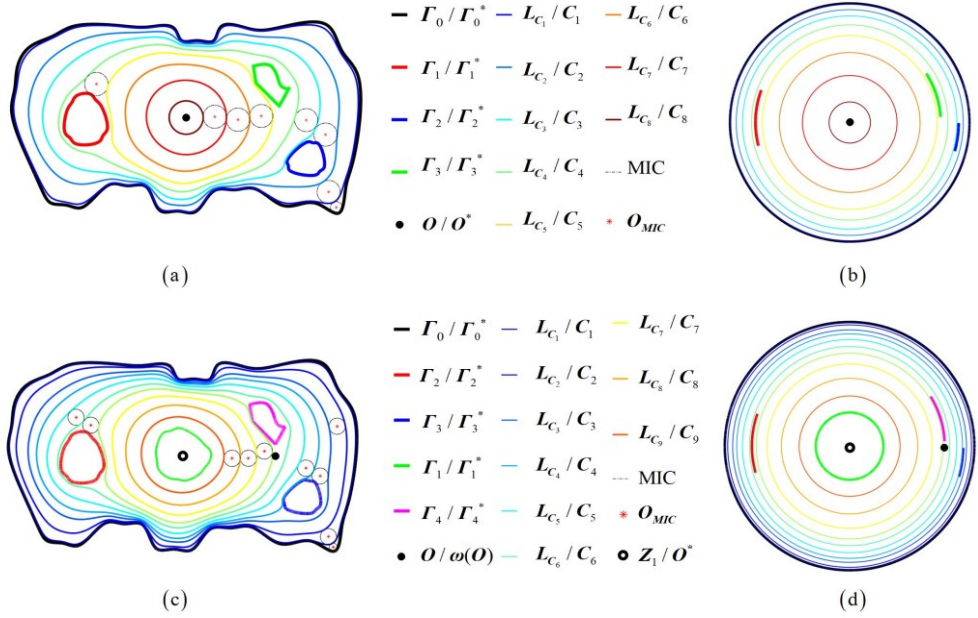

**Figure 5.** The iso-parameters within the preimage and mapping domains, the symbols on the left side of / correspond to the curve or points in the left figure, while the symbols on the right side of / correspond to the curve or points in the right figure.

Let $R_{C_0} = 1$, which indicates that $L_{C_0}$ is the outer boundary $\Gamma_0$. We want to find a set of decreasing radii $\{R_{C_1}, R_{C_2}, \ldots R_{C_k}\}$, $1 > R_{C_1} > R_{C_2} > , \ldots, > R_{C_k}$ to make $L_{C_0-C_1} = \frac{C}{2}$, $L_{C_i-C_{i+1}} = C, i = 1,2, \ldots, k-1$, and $0 < L_{C_k-C_k} \leq C$, where $C$ is a constant. When $R_i$ is determined, $L_{C_i-C_{i+1}}$ decreases monotonically with $R_{i+1}$. Thus, if $R_{C_i}$ is determined, $R_{C_{i+1}}$ can be solved by dichotomy between $[0, R_{C_i}]$ for case I and between $[R_1, R_{C_i}]$ for case II, and when $0 \leq L_{C_i-C_i} < C$ is met, $i$ is denoted as $k$, the operation completes, and the radius solution stops. The maximum inscribed circle of $G_{C_i} \cap C_{G_{C_j}} \cap G$ or $G_C \cap G$ can be calculated based on Delaunay triangulation (Birdal, 2023). Let $P$ denote the discrete points on the boundary of domain $B$ where the maximum inscribed circle needs to be found. $P$ is connected to triangle mesh $T$ by Delaunay triangulation with $k$ triangular patches $T_1, T_2, \ldots, T_k$, and the center and the radii of the circumferential circle of each triangular patch are denoted as $OT_1, OT_2, ..., OT_k$ and $RT_1, RT_2, ..., RT_k$, respectively. The maximum inscribed circle $(O_{MIC}, R_{MIC})$ can be represented as:

$$(O_{MIC}, R_{MIC}) \ for \ R_{MIC} = max\{RT \ |OT \in \{OT_1, OT_2, ..., OT_k\}, OT \in B\} \tag{1}$$

Figure 6 shows the first dichotomous iterative solution to the maximum inscribed circle in Figure 5(a), where $R_{C_0} = 1$, and $R_{C_1}$ iterates to 0.5. We pick out the triangular patch as the green patch with the circumcircle center on $B$ and finds the largest circumcircle radius.

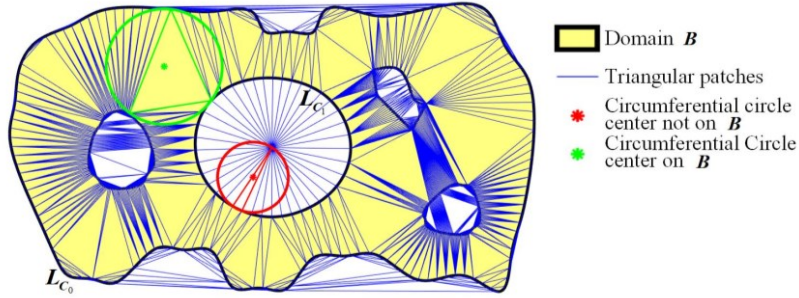

**Figure 6.** Finding the maximum inscribed circle by Delaunay triangulation.

For convenience, we denote the pre-image of a circle with a radius of $R_x$ in the mapping domain as $L_{R_x}$, expressed as $L_{R_x} = \omega^{-1}(R_x)$. The pseudocode for determining iso-parameters $L_{C_1}, L_{C_2}, ..., L_{C_k}$ with controlled spacing between $L_{C_i}$ and $L_{C_{i+1}}$, and their corresponding circle radii $R_{C_1}, R_{C_2}, ..., R_{C_k}$ in the image domain is presented below:

---
**Algorithm:** Calculate iso-parameters with a suitable adjacent spacing.
    **Input:** Maximum inscribed circle radius set between adjacent iso-parameters $C$, allowable calculation error for the maximum inscribed circle between adjacent iso-parameters $\varepsilon$, The radii of the boundary $\Gamma$ in the image domain $R_0, R_1, ..., R_m$.
    **Output:** $Lk = \{L_{C_1}, L_{C_2}, ..., L_{C_k}\}$ and $Rk = \{R_{C_1}, R_{C_2}, ..., R_{C_k}\}$.
1   $R_{up} = 1; R_{down} = 0$ for case I, $R_{down} = R_1$ for case II; $k = 0$; $Rk = \{\}$; $Lk = \{\}$;
2   Making $R_0, R_1, ..., R_m$ and $R_{down}$ global variables allow them to be accessed in the subroutine $MIC()$;
3   **While** 1 **do**
4      $R_{MIC} = C$;
5      **if** $k = 0$ **then**
6         $R_{MIC} = \frac{C}{2}$;
7      **end**

| 8 | $L_{R_{up}-R_{up}} = MIC(R_{up}, R_{up})$; |
| 9 | **if** $L_{R_{up}-R_{up}} < R_{MIC}$; **then** |
| 10 |    BREAK; |
| 11 | **end** |
| 12 | $R_{up\_t} = R_{up}$; $R_{down\_t} = R_{down}$; $k = k + 1$; |
| 13 | **While** 1 **do** |
| 14 |    $R_k = \dfrac{R_{up\_t} + R_{down\_t}}{2}$; |
| 15 |    $L_{R_{up}-R_k} = MIC(R_{up}, R_k)$; |
| 16 |    **if** $\left|L_{R_{up}-R_k} - R_{MIC}\right| < \varepsilon R_{MIC}$ **then** |
| 17 |      BREAK; |
| 18 |    **end** |
| 19 |    **if** $L_{R_{up}-R_k} > R_{MIC}$ **then** |
| 20 |      $R_{down\_t} = R_k$; |
| 21 |    **else** |
| 22 |      $R_{up\_t} = R_k$; |
| 23 |    **end** |
| 24 | **end** |
| 25 | $R_{up} = R_k$; |
| 26 | $Rk = \{Rk, R_k\}$; |
| 27 | $L_{C_k} = \omega^{-1}(R_k)$; $Lk = \{Lk, L_{C_k}\}$; |
| 28 | **end** |
| 29 | Return $Rk, Lk$; |

where $MIC(R_a, R_b)$ is the function used to calculate the maximum inscribed circle $R_{mic}$ within the domain $G_{R_a} \cap C_{G_{R_b}} \cap G$ or the domain $G_{R_a} \cap G$, denoted uniformly as $G_M$. The pseudocode for this function is as follows:

**Algorithm** $MIC$: Calculate the maximum inscribed circle for the multi-connected domain between two adjacent iso-parameters $G_M$.

    **Input:** Upper limit of the iso-parameter radius $R_a$, lower limit of the iso-parameter radius $R_b$.
    **Output:** Maximum inscribed circle radius $R_{mic}$ of $G_M$ between $L_{R_a}$ and $L_{R_b}$.

| 1 | **if** $R_a == R_b$ |
| 2 |    $R_b = R_{down}$; |
| 3 | **end** |
| 4 | $L_{R_a} = \omega^{-1}(R_a)$; $L_{R_b} = \omega^{-1}(R_b)$; |
| 5 | $G_M = \{L_{R_a}, L_{R_b}\}$; |
| 6 | **for** $i = 0: m$ |
| 7 |    **if** $R_a > R_i > R_b$ |
| 8 |      Sequentially add $\Gamma_i$ to the boundary of $G_M$, $G_M = \{G_M, \Gamma_i\}$; |
| 9 |    **end** |
| 10 | **end** |
| 11 | Transform the set of points on all boundaries of $G_M$ into a triangular mesh through Delaunay triangulation; |

| 12 | Calculate the circumcenter $\{OT_1, OT_2, ..., OT_k\}$ and radius $\{RT_1, RT_2, ..., RT_k\}$ for each triangle facet; |
|---|---|
| 13 | Determine whether the circumcenter of each triangle facet is inside the domain enclosed by the first boundary $G_M\{1\}$ of $G_M$. If not, remove this triangle facet from the triangular mesh; |
| 14 | **for** $i = 2: length(G_M)$ |
| 15 |     Determine whether the circumcenter $OT_1, OT_2 ...$ of each remaining triangle facet is inside the polygonal domain enclosed by the $i$ - th boundary of $G_M$. If so, remove this triangle facet from the triangular mesh; |
| 16 | **end** |
| 17 | Find the maximum circumcircle radius $R_{mic}$ among the remaining triangular facets; |
| 18 | Return $R_{mic}$; |

2.2.2 Generate spiral path based on iso-parameters spiral offsetting

We believe that the iso-parameters mentioned in Section 2.2.1 can be used for cavity machining, but the iso-parameters need to be bridged. To eliminate bridging and make the path smoother, in this section, spiral paths that do not pass through the hole and without bridging will be generated based on the calculated $\{R_{C_0}, R_{C_1}, ..., R_{C_k}\}$ for iso-parameters.

In the plane with the Cartesian coordinate system, a point set $SP$ is defined as follows:

$$SP = \begin{bmatrix} X_{SP} \\ Y_{SP} \end{bmatrix} = \begin{bmatrix} \theta_0 & \theta_0 & \theta_1 & \cdots & \theta_k & \theta_{k+1} & \theta_{k+1} \\ R_{C_0} & R_{C_0} & R_{C_1} & \cdots & R_{C_k} & R_{min} & R_{min} \end{bmatrix} \quad (2)$$

where $R_{min} = 0$ for case I, $R_{min} = R_1$ for case II. $\theta_0, \theta_1 ... \theta_{k+1}$ are pending parameters. $SP$ is shown in Figure 7(a)(b) as black points. The first two points $SP[:,1,2]$ and the last two points $SP[:, end - 1, end]$ are coincide with the start and end points of the spiral curve, respectively. If we want a spiral path that moves clockwise from inside to outside with a maximum spacing almost identical to $C$, the pending parameters are subject to the following constraints:

$$\begin{aligned} &\theta_0 < \theta_1 < \theta_2 < \cdots < \theta_k \leq \theta_{k+1} \\ &|\theta_{j+1} - \theta_j| \geq 2\pi \quad j = 0,1,2, ..., k \end{aligned} \quad (3)$$

In Eq. 3, the first constraint guarantees the direction of rotation and the second constraint is to avoid excessive spacing between the spiral itself. As shown in Figure 7(a)(b), the circular slit is also mapped in the form of polar coordinates to the plane where the $SP$ located, and their image is repeated in $2\pi$ cycles along the $X$-axis. We connect $SP$ with smooth curves $SC$ and chooses the cubic B-spline curve to connect. By adjusting $\theta_0, \theta_1, \ldots, \theta_{k+1}$, $SC$ can be made not intersect with the images of slits in the plane. This seems to have many parameters to adjust, but the following restrictions can be made to meet the requirement of most scenarios by adjusting one parameter $\theta_0$:

$$\begin{cases} -\pi < \theta_0 \leq \pi \\ \theta_{j+1} - \theta_j = 2\pi \quad j = 0,1,\ldots,k \end{cases} \quad (4)$$

where $\theta_0$ can make curve $SC$ pan along the $X$-axis between the two blue dashed lines in Figure 7(a)(b), thereby determining the two end points of the paths.

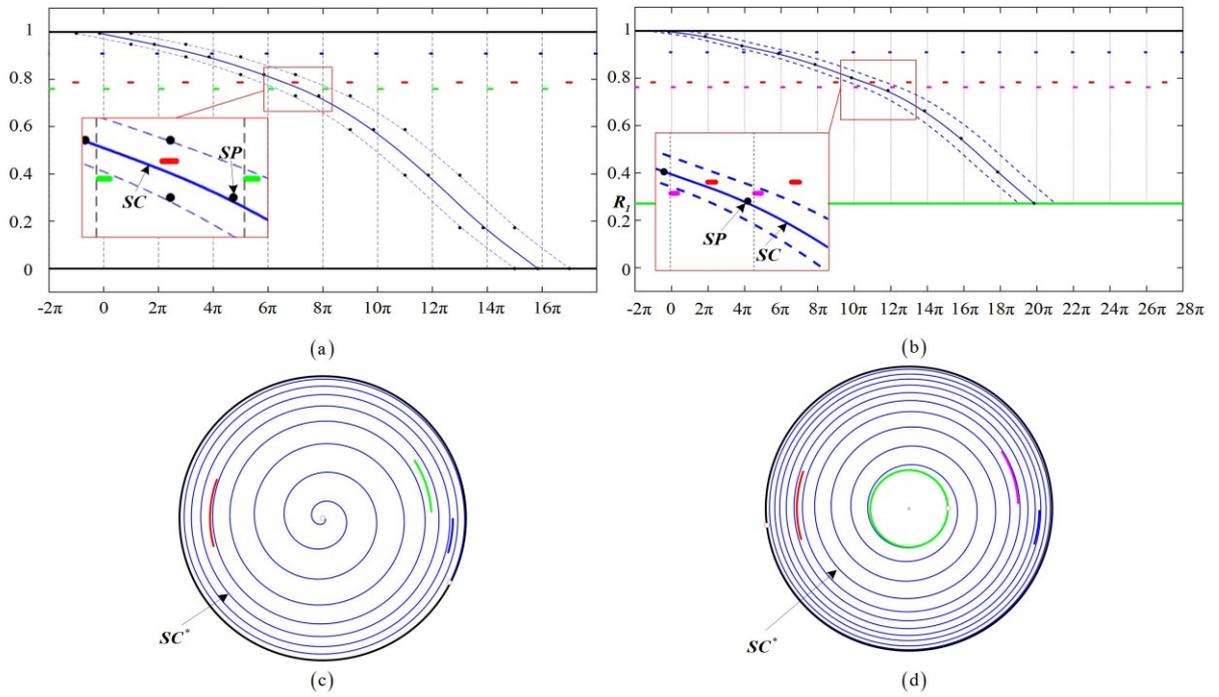

**Figure 7.** Generating spirals that do not intersect slits by transforms curves in Cartesian coordinates to polar coordinates.

The Cartesian coordinate values of the points on $SC$ are converted to the polar coordinate values by:

$$\begin{cases} \rho_{SC} = X_{SC} \\ \theta_{SC} = Y_{SC} \end{cases} \quad (5)$$

In this way, an unequal spiral curve $SC^*(\rho_{SC}, \theta_{SC})$ on $D$ or $A$ can be obtained, as the blue solid thin lines in Figure 7(a)(c); then, the spiral trajectory $SC^{-1}$ in domain $G$ can be obtained by $SC^{-1} = \omega^{-1}(SC^*)$, as the blue thin lines shown in Figure 8. As shown by the blue dashed thin lines in Figure 8, at the beginning and end of the path, a part of the curve does not affect whether the path can be completely covered, so the envelope method (Shen et al., 2022) can be used to identify curve segments that do not affect coverage, as the local enlarged drawing shown in Figure 8(a), and we only adjust the start and end of the path from red solid points to red hollow points. The blue solid lines in Figure 8 illustrate the generated spiral paths.

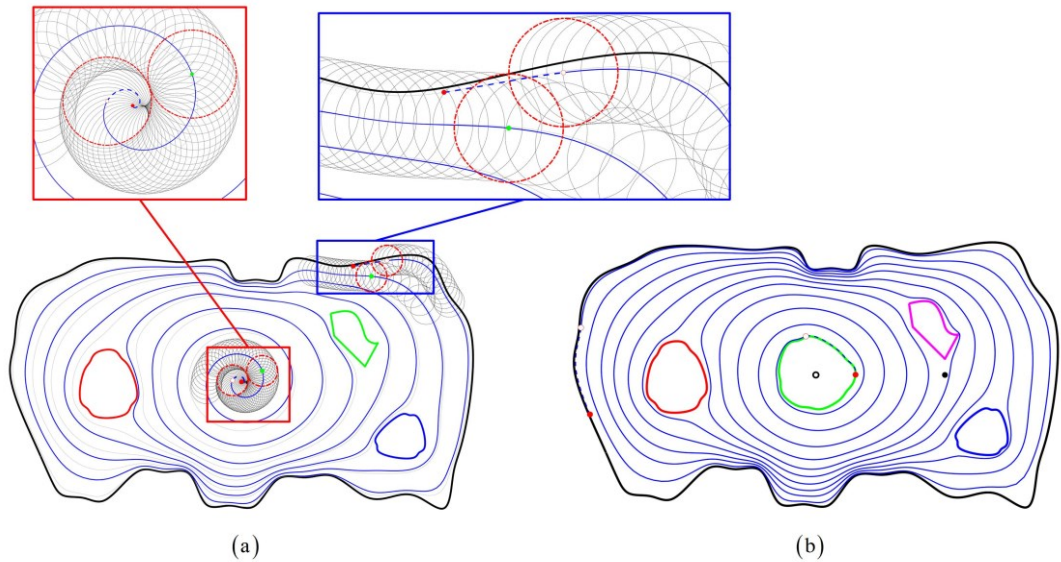

**Figure 8.** The spiral trajectories generated on multi-connected domains.

The spiral trajectory generated by our algorithm and the traditional PDE-based method is boundary-adapted, so it can be easy to find two transition points on the boundary $\Gamma_j$ and spiral trajectory $SC^{-1}$ that are close and have an approximately normal vector direction. Around these two points, $\Gamma_j$ and $SC^{-1}$ can be fused into a trajectory with an intersection point through a rollercoaster-like rotation, as illustrated in local enlarged drawing in Figure 1. This ensures the trajectory does not produce violent turns when transitioning to the boundary path. Note that the iso-parameter spacing control and offsetting strategies used in examples in Figure 1(a) and Figure 1(b) are the same as we mentioned in Section 2.2.1 and Section 2.2.2, and the maximum path spacing limit is set to 12 mm.

*2.3 Analysis of algorithm complexity*

In Section 2.1, the complexity of calculating CSM on boundary $\omega(\Gamma)$ is $O\big((m+1)n\log(n)\big)$, and the complexity of using Cauchy interpolation to find positive mapping $\omega(\hat{n})$ and inverse mapping $\omega^{-1}(\hat{n})$ for an internal point is $O\big((m+1)n+\hat{n}\big)$, where $m$ denotes the number of boundaries, $n$ denotes the number of discrete points per boundary, $\hat{n}$ denotes the number of points to be solved by Cauchy interpolation (Nasser, 2015).

In Section 2.2.1 and Section 2.2.2, for our SCCP generation method, the highest complexity is caused by determining $\{R_{C_0}, R_{C_1}, \dots R_{C_k}\}$ for iso-parameters. Thus, the algorithm complexity analysis elsewhere in Section 2.2.1 and Section 2.2.2 is omitted. Firstly, as mentioned before, if $R_i$ is calculated, $R_{i+1}$ can be calculated by dichotomy with multiplied complexity $O\left(\log\left(\frac{1}{\varepsilon}\right)\right)$, where $\varepsilon$ represents the allowable dichotomy iteration tolerance ratio of the maximum inscribed circle radius. In each dichotomous iteration, it is required to solve the maximum inscribed circle on a multi-connected domain $B = G_{C_i} \cap C_{G_{C_j}} \cap G$ or $B = G_C \cap G$ once, and the multiplied complexity is $O\big(Pt_{\partial B}\log(Pt_{\partial B})\big)$ (Birdal, 2023), where $Pt_{\partial B}$ denotes the total number of discrete points on boundary $\partial B$. In the worst case as shown in Figure 6, $Pt_{\partial B} = (m+1)n + \hat{n}$, so the upper multiplied complexity of finding the maximum inscribed circle radius in each dichotomy process is $O\left(\big((m+1)n+\hat{n}\big)\log\big((m+1)n+\hat{n}\big)\right)$. The above operation is repeated $k$ times. Thus, the worst-case complexity is:

$$O\left(k\log\left(\frac{1}{\varepsilon}\right)\big((m+1)n+\hat{n}\big)\log\big((m+1)n+\hat{n}\big)\right) \tag{7}$$

The complexity of the algorithm in determining $\{R_{C_0}, R_{C_1}, \dots R_{C_k}\}$ is higher than that in Section

2.1, so the complexity of the whole algorithm is subject to Eq. 7.

## 3. Experiment and analysis

*3.1 Numerical arithmetic experiments*

Several typical test cases show in Figure 9 are conducted to verify the versatility of our algorithm in different multi-connected regions. The calculation time of the two use cases in Figures 9(a)(b) shows that the number of path rotation turns $k$ and the spend times are linear, and the operation cost that increases with $k$ linearity is acceptable for offline cavity machining path planning. Figures 9(d)(e) are multi-connected regions with ruffle shaped boundaries. If a single-connected subregion division method is adopted, the number of subregions divided by these two use cases will be large and require a large number of bridges. Figure 9(d) shows that the strip island around the origin has little effect on the uniformity of the path spacing. But Figure 9(g) shows that the strip island pointing to the origin significantly reduces the uniformity of the trajectory spacing, which is the limitation of our algorithm and needs further investigation. Figure 9(f) shows the potential of our algorithm to be combined with current subregion division methods.

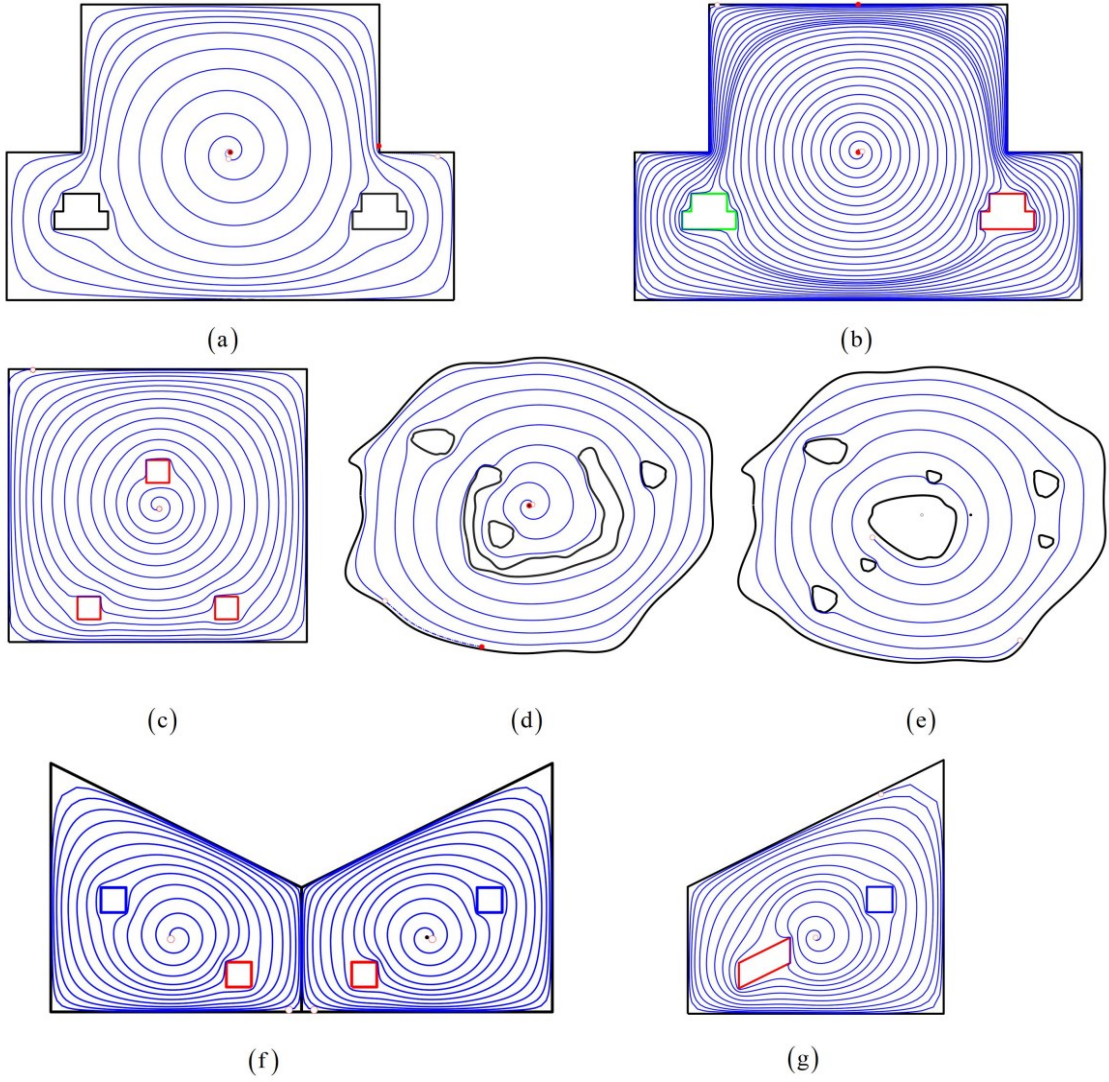

**Figure 9.** The generated spiral paths in different multi-connected domains.

Table 1 lists the set parameters and time taken to run our algorithm on a computer equipped with an Intel i5-10400F CPU and an NVIDIA GeForce GTX-1600 GPU. We calculate the ratio of predict operation number and the calculation spend time $STR$ by:

$$STR = \frac{klog\left(\frac{1}{\varepsilon}\right)((m+1)n+\hat{n})log((m+1)n+\hat{n})}{\text{Spend Times}} \tag{8}$$

$STR$ are normalized to obtain $\widetilde{STR}$ as $\widetilde{STR} = \frac{STR}{\max(STR)}$, and the normalized values $\widetilde{STR}$ are presented in a box plot in Figure 10. From Figure 10, it can be seen that the value of $\widetilde{STR}$ concentrate between 0.5 and 1, indicating that the predicted computational complexity aligns with the actual runtime without significant order-of-magnitude fluctuations, validating the effectiveness of our complexity analysis.

**Table 1.** The parameters and spend time of the test cases.

| Test Cases | $m$ | $n$ | $\hat{n}$ | $k$ | $\varepsilon$ | Spend Time (s) | $STR$ |
|---|---|---|---|---|---|---|---|
| Figure 8(a) | 3 | 8192 | 1000 | 8 | 0.01 | 290.04 | 44725.34 |
| Figure 8(b) | 4 | 8192 | 1000 | 10 | 0.01 | 442.90 | 46440.79 |
| Figure 9(a) | 2 | 8192 | 1000 | 11 | 0.01 | 267.09 | 49232.75 |
| Figure 9(b) | 2 | 8192 | 1000 | 29 | 0.01 | 683.25 | 50738.47 |
| Figure 9(c) | 3 | 2048 | 1000 | 18 | 0.01 | 172.71 | 40262.01 |
| Figure 9(d) | 4 | 2048 | 1000 | 9 | 0.01 | 85.63 | 50743.62 |
| Figure 9(e) | 7 | 2048 | 1000 | 7 | 0.01 | 92.17 | 59360.93 |
| Figure 9(f) (left part) | 2 | 8192 | 1000 | 14 | 0.01 | 288.75 | 57959.57 |
| Figure 9(g) | 2 | 4096 | 1000 | 15 | 0.01 | 116.14 | 75039.88 |
| Figure 1(a) | 3 | 4096 | 1000 | 20 | 0.01 | 231.43 | 67546.46 |
| Figure 1(b) | Control experiments | | | | | 82.01 | |

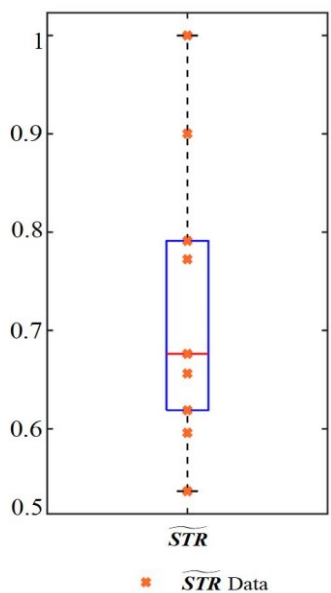

**Figure 10**. The box plot of $\widetilde{STR}$.

*3.2 Milling experiment on path coverage and machining time consumption*

In this section, a robot cavity milling comparative experiment is conducted to demonstrate the milling time, cavity shape and milling residual areas between the CSM-based milling paths and the PDE-based milling path (Chuang & Yang, 2007; Driscoll, 2023), which is a classical spiral generation algorithm that has been widely used in cavity milling. Figure 11 shows the structure of the milling test bench equipped with a robot arm model IRB-6700, workpieces. The tool diameter $D$ is equal to the maximum paths spacing with $D$ = 12mm. The feed speed $vf$ is set to 40mm/s.

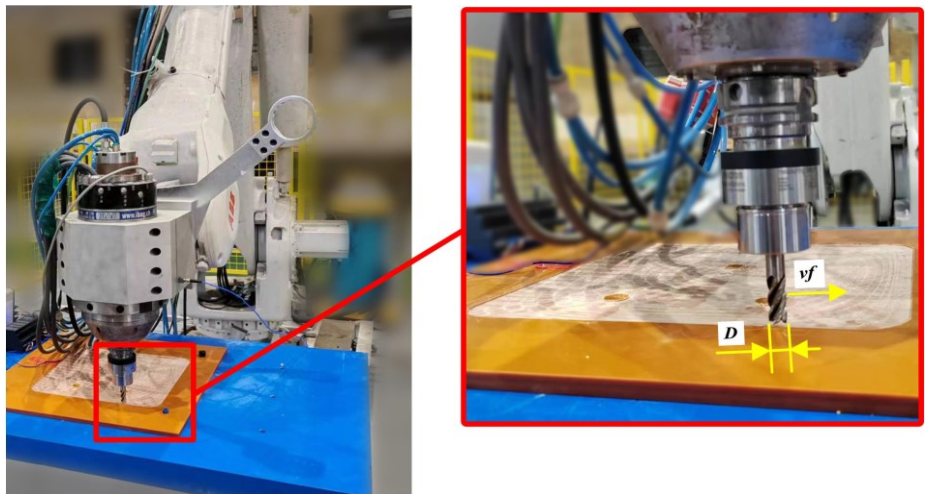

**Figure 11.** The test bench and workpiece.

Figure 12 shows the shape of the two workpieces milled by CSM-based paths and the PDE-based path, respectively. CSM-based path reduced path length by 12.70% compared with PDE-based path (12581.84 mm for CSM-based milling paths vs 14412.83 mm for PDE-based path) and the CSM-based path reduced milling time by 12.33% compared with PDE-based path (316.46 sec for CSM-based milling paths vs 360.99 sec for PDE-based path), the machining duration for fixed trajectories typically fluctuates by less than 2%, demonstrate the practical value of algorithms to improve cavity milling efficiency. It can be observed that all workpieces are machined with clear cavity boundaries and no residues inside the cavity or scallop-shaped residues near the boundaries. Thus, after fusing the boundaries of the multi-connected region as a part of trajectory into the spiral path, the fused path achieves the complete coverage required for machining the cavity. The widest part of the milling pattern is 12 mm, which indicates that the upper limit spacing of the path is controlled to be equal to the tool diameter $D$. Thus, based on the specified tool diameter, our spiral trajectory generation strategy results in the widest spacing of complete coverage paths, which also means the paths are less redundant coverage and shorter.

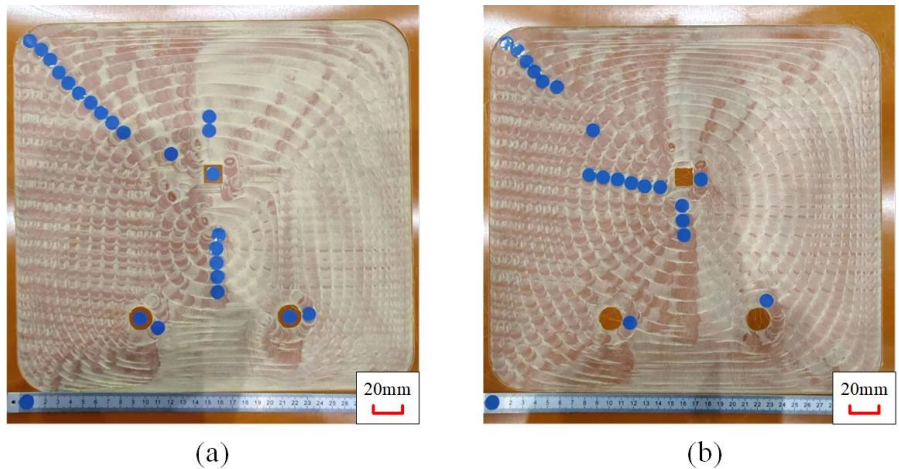

(a)                 (b)

**Figure 12.** The final morphology of the workpiece from two sets of experiments. (a) Cavity 1 machined with the CSM-based path with path length 12581.84mm. (b) Cavity 2 machined with the traditional PDE-base path with path length 14412.83mm.

*3.3 Dry run experiment on trajectory steering impact*

Compared with milling processes, dry run experiments more accurately capture the influence of trajectory smoothness on the inertial shocks experienced by the actuator. Inertial shocks are pivotal in non-contact processing, necessitating the actuator to restrict its operating speed to mitigate inaccuracies resulting from these shocks(Yang et al., 2002). Dry run experiment results demonstrate higher signal-to-noise ratios compared with milling experiments, enhancing their reliability. This can be attributed to the reduced impact of factors such as robot positioning accuracy, milling force measurement precision, and uncertainties arising from the complex cutting mechanisms during milling processes on dry run experiments.

To precisely assess the differences in trajectory smoothness between the two paths, a dry run experiment is conducted: the milling tool is take off the robot, allowing it to alternate between Path 1 and Path 2 at a speed of 200 mm/s. After each run along a path, the robot pause for 5 seconds. Triaxial accelerometer is attached with adhesive to the robot's end effector spindle to collect acceleration signals from the actuators, as illustrated in Figure 13. The accelerometer sensor model used is PCB-356A44, with the sampling frequency equals to 2000Hz. The total acceleration $a_T$ is calculated based on the acceleration signals collected from the three channels $(X,Y,Z)$ using the following formula:

$$a_T = \sqrt{a_X^2 + a_Y^2 + a_Z^2} \tag{9}$$

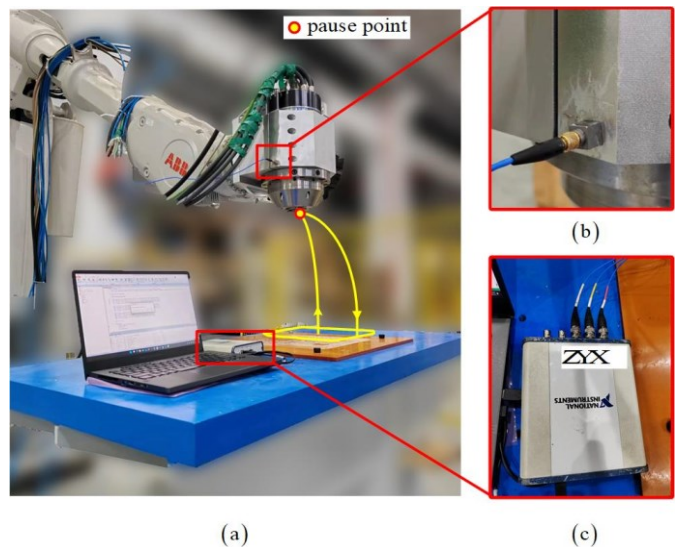

**Figure 13.** The test bench and workpiece. (a) Experimental platform. (b) Triaxial accelerometer affixed to the spindle. (c) Signal acquisition card.

Figure 14(a) displays the signals collected as the robot alternately traverses Path 1 and Path 2 for 10 cycles. As depicted in Figure 14(a), the acceleration measurement results for each experiment remain relatively stable during dry runs, indicating the high reliability of the collected data. Figure 14(b) presents the acceleration signals for the second cycle. Clearly visible in the figure is the impact on the actuator caused by the cutting in and lifting of the tool due to the non-smooth nature of the trajectory. These impacts are undesirable and should be minimized during the machining process. In contrast, the spiral trajectory exhibits only one instance of cutting in and lifting of the tool compared with other trajectories.

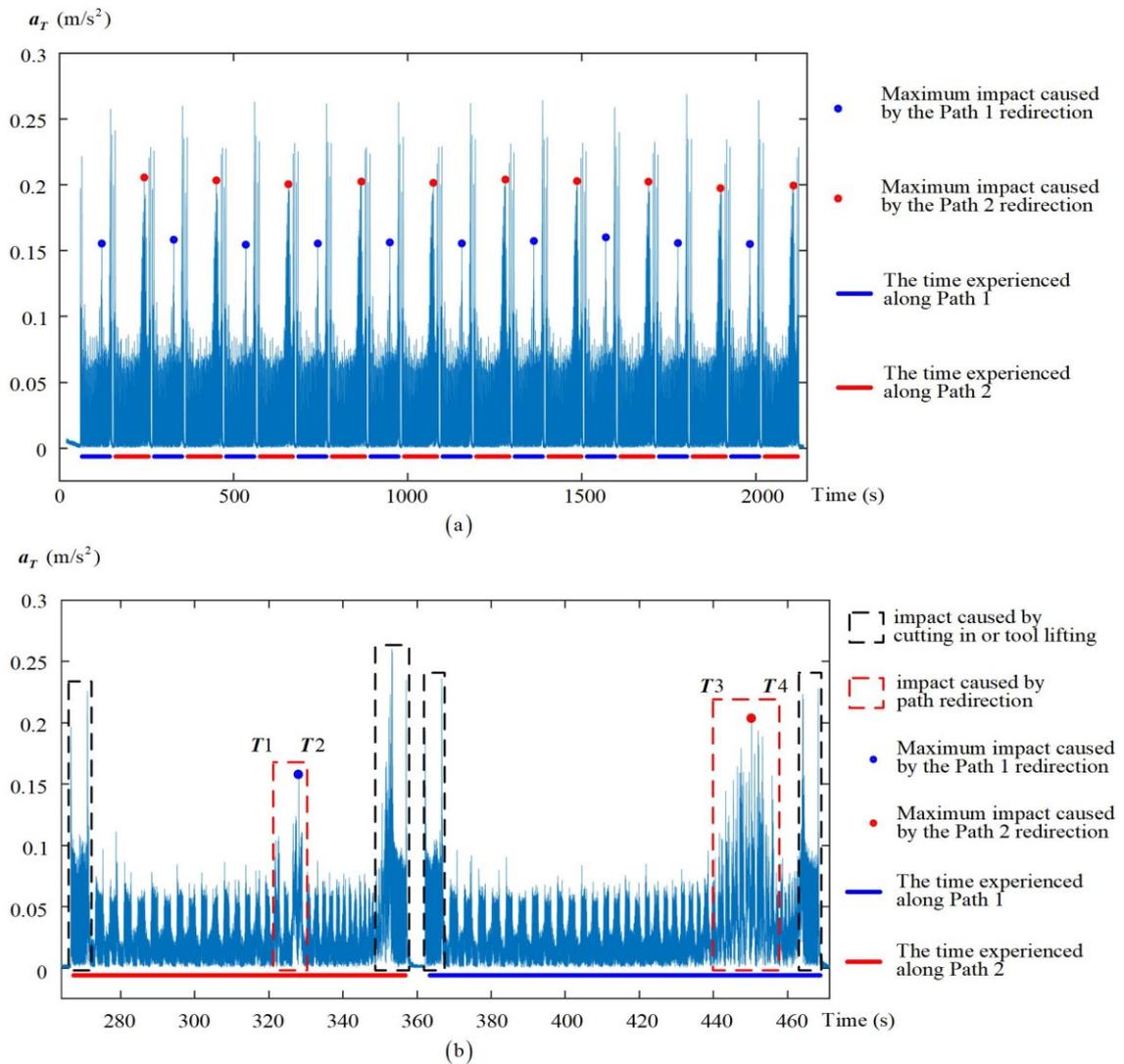

**Figure 14.** The acceleration collected by using accelerometer sensor alternately ran along

Path 1 and Path 2 for 10 cycles. (a) Overall data for the 10 cycles. (b) Data for the second

cycle.

Excluding the significant accelerations resulting from cutting in and lifting the tool, the remaining acceleration is attributed to the sharp turns in the trajectory. Figure 14(b) illustrates that, in comparison to Path 1, Path 2 shows an increase in both the intensity and duration of impacts. We deduce the impact positions on the path by calculating the times ($T1$~$T2$ for Path 1, $T3$~$T4$ for Path 2) of force initiation and termination, taking into account the velocity of the actuator. The corresponding paths are highlighted in red in Figure 15.

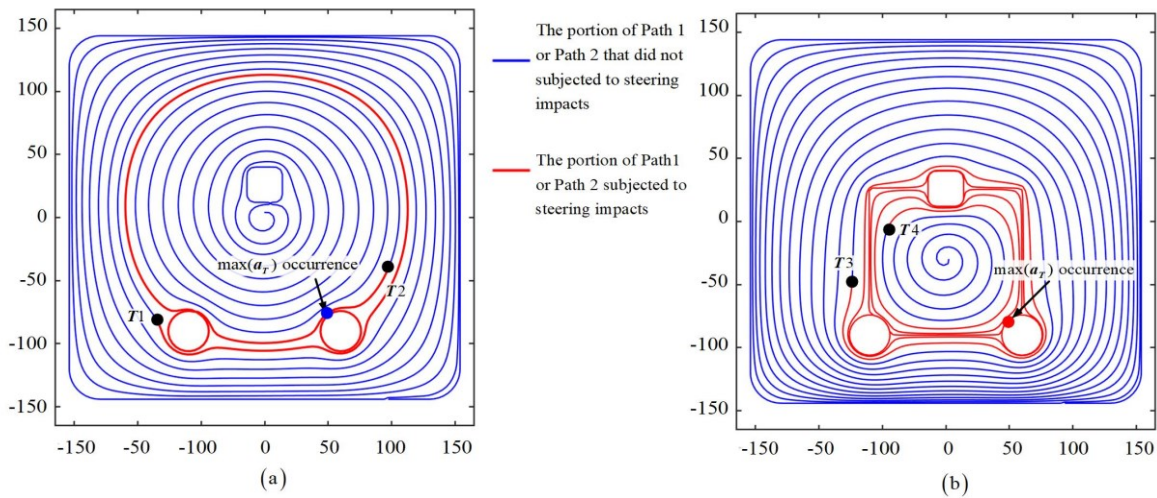

**Figure 15.** Correspondence between Trajectories and Impact Locations.

In Figure 15, it is evident that the impact of Path 2 initiates at the transition point where the trajectory shifts towards the inner side of the square region. In this region, the trajectory becomes less smooth due to the influence of the boundary of the subregion, while the turning of Path 1 remains relatively gentle.

Figure 16 presents box plots of the maximum impact forces induced by the path collected over 10 cycles. In the dry run experiment, the average maximum impact on Path 1 decreased by 22.78% compared with the average maximum impact on Path 2 ($0.1529 mm/s^2$ for Path 1 vs $0.1980 mm/s^2$ for Path 2). It is evident that the upper and lower limits of the two box plots do not overlap, confirming that Path 1 experiences smaller impacts compared with Path 2. The experimental results showcase the powerful potential application of our algorithm in non-contact high-speed machining scenarios, such as 3D printing (Cam & Gunpinar, 2023).

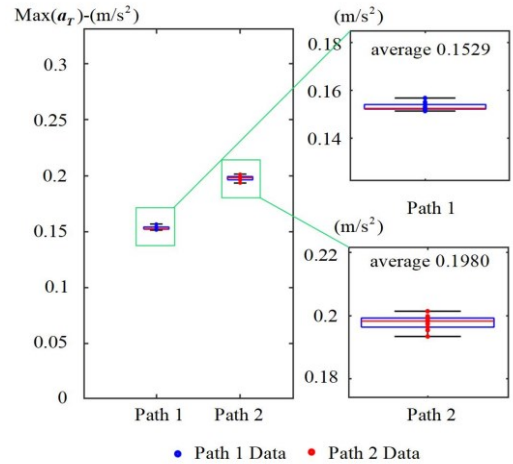

**Figure 16.** Box plots of the maximum impact forces induced by different paths collected over 10 cycles.

## 4. Summary and outlook

*4.1 Summary*

We propose an offline CSM-based spiral path planning method in the multi-connected region for cavity milling, which is based on the proposed CSM iso-parameters spacing control strategy and CSM iso-parameter spiral offsetting strategy. Compared with traditional spiral path planning methods, the new method does not divide subregions, therefore, a smoother and more continuously connected spiral trajectory can be generated to cover the entire region with fewer discontinuities. This shortens the length, improves the smoothness, and enhances the uniformity in path spacing of spiral path. The upper limit for spiral path spacing can be set, which allow us to select the widest and shortest spiral path according to the milling cutter diameter. The spiral only path is not a complete coverage trajectory, but we propose a strategy to fuse the boundary curves and the spiral trajectory into one path, and the fused path is the complete coverage path.

Through multiple numerical arithmetic experiments, we demonstrate the strong applicability of the proposed spiral path planning method under complex boundary conditions and its potential for integration with other subregion division methods. But in domains with strip islands pointing toward the origin, the path spacing generated by our method is incompetent and need further research.

The upper bound of algorithm complexity is $O\left(klog\left(\frac{1}{\varepsilon}\right)((m+1)n+\hat{n})log((m+1)n+\hat{n})\right)$, and the computational overhead is directly proportional to the number of spiral rotations number $k$, which is considered acceptable in offline path generation of cavity milling.

Based on the final shapes achieved by using two different paths for milling two cavities, it is evident that our algorithm fulfills the requirement for complete coverage. In comparison to traditional PDE-based methods, our algorithm has reduced the path length by 12.70% and machining time by 12.33%, showcasing the practical value of the algorithm in enhancing cavity milling efficiency. In the dry run experiment, our algorithm decreased the inertial impact by 22.78% compared with the traditional PDE-based algorithm, enabling the actuator to accomplish complete coverage in less time while minimizing impact.

*4.2 Outlook*

We will explore the potential of our path-planning algorithm in diverse fields, including UAV rescue operations. Extending our algorithm to the problem of complete coverage of three-dimensional surfaces is also worthy of further discussion.

**Acknowledgments**

This work was supported by the National Natural Science Foundation of China (52188102，52375495). The authors thanked M. M. S. Nasser for the key guidance on conformal slit mapping. The authors are also very grateful to Gary P. T. Choi., Wu Kang and Jean. Feydy for their invaluable guidance in homomorphic mapping. We would like to thank Dr. Michael Otte for his meticulous review and recommendation of the paper, and thank the two anonymous reviewers for their valuable comments.

**Funding**

This research received no specific grant from any funding agency in the public, commercial, or not-for-profit sectors.

**Declaration of conflicting interests**

The Authors declare that there is no conflict of interest.

# Appendix

Appendix describes the implementation of these two special types of slit mappings and their inverse mappings used in our paper.

A-1. Boundary curve parameterization

As shown in Figure 2, supposing $G$ is a bounded open $m$-multiply ($m > 1$) connected domain with origin $O \in G$ in the extended complex plane $C \cup \{\infty\}$ and has $(m + 1)$-bounds $\partial G = \Gamma = \Gamma_0 \cup \Gamma_1, \ldots, \Gamma_{m-1} \cup \Gamma_m$, we denote $G \cup \partial G$ as $\bar{G}$, where $\Gamma_0, \Gamma_1, \ldots, \Gamma_m$ are closed Jordan curves, and $\Gamma_j$, $j = 0,1,\ldots,m$ can be smooth or segmentally smooth with $p_j$ corner points. Meanwhile, we denote the outer curve encloses the origin as $\Gamma_0$ and all the other inner curves as $\Gamma_1, \Gamma_2, \ldots, \Gamma_m$. Besides, we suppose that the direction of $\Gamma_0$ is counterclockwise, and the direction of $\Gamma_1, \Gamma_2, \ldots, \Gamma_m$ is clockwise. In annular slit mapping, except for the origin, a point $Z_1$ enclosed by an inner curve needs to be specified, and we denote the inner curve encloses $Z_1$ as $\Gamma_1$.

In practical engineering applications, boundary curves usually do not have an explicit expression, but boundary curves can be fit with a combination of arcs, straight segments, and splines. For easier practical use, based on the Nyström method, we introduce cubic B-spline for boundary curve parameterization. The cubic B-spline $P(t)$ has a minimum of 4 control points, it can be represented as:

$$\begin{cases} P(t) = \sum_{j=0}^{k-4} \sum_{i=0}^{3} P_{j+i}\, F_{i,3}(t-j)\delta(t-j) & t \in [0,\ k-3]\ for\ open\ curve \\ P(t) = \sum_{j=0}^{k-1} \sum_{i=0}^{3} P_{j+i-k floor\left(\frac{j+i}{k}\right)}\, F_{i,3}(t-j)\delta(t-j)\ t \in [0,\ k]\ for\ closed\ curve \end{cases} \quad (A\text{-}1)$$

where $t$ denotes the curve parameter, $P_0, P_1, \ldots, P_{k-1}, k \geq 4$ denote the control points, $floor(\cdot)$ represents the rounding-down operator, $F_{i,3}\ i = 1,2,3$ denote the cubic B-spline basis function in the following form:

$$\begin{cases} F_{0,3}(t) = \frac{1}{6}(1-t)^3 \\ F_{1,3}(t) = \frac{1}{6}(3t^3 - 6t^2 + 4) \\ F_{2,3}(t) = \frac{1}{6}(-3t^3 + 3t^2 + 3t + 1) \\ F_{3,3}(t) = \frac{1}{6}t^3 \end{cases} \quad \text{(A-2)}$$

$\delta(t-j)$ has the following form:

$$\delta(t-j) = \begin{cases} 1 & 0 \leq t-j \leq 1 \\ 0 & other\ cases \end{cases} \quad \text{(A-3)}$$

The first derivative of $P(t)$ has the following form:

$$\begin{cases} P'(t) = \sum_{j=0}^{k-4} \sum_{i=0}^{3} P_{j+i}\, F'_{i,3}(t-j)\delta(t-j)\ t \in [0,\ k-3] & for\ open\ curve \\ P'(t) = \sum_{j=0}^{k-1} \sum_{i=0}^{3} P_{j+i-k\,floor\left(\frac{j+i}{k}\right)}\, F'_{i,3}(t-j)\delta(t-j)\ t \in [0,\ k] & for\ closed\ curve \end{cases} \quad \text{(A-4)}$$

where:

$$\begin{cases} F'_{0,3}(t) = -\frac{1}{2}(1-t)^2 \\ F'_{1,3}(t) = \frac{3t^2}{2} - 2t \\ F'_{2,3}(t) = -\frac{3t^2}{2} + t + \frac{1}{2} \\ F'_{3,3}(t) = \frac{1}{2}t^2 \end{cases} \quad \text{(A-5)}$$

If $p_j = 0\ j \in \{0,1, \dots, m\}$, closed boundary curve $\Gamma_j$ is smooth without any corners, and it can be fitted to a closed cubic B-spline $\eta_j(t)$ with $k$ noncoincidence control points and curve parameter $t \in [0, k]$. We use $\hat{t}$ instead of $t$ as the curve parameter as the form $\eta_j(t(\hat{t}))$ and $t(\hat{t})$ can be represented by:

$$t(\hat{t}) = \frac{\hat{t}k}{2\pi}\ \hat{t} \in [0, 2\pi]\ for\ closed\ curve \quad \text{(A-6)}$$

By converting the algebraic symbol $\hat{t}$ to $t$, we have $\eta_j(t)\ t \in J_j = [0, 2\pi]$. Thus, we obtain the $2\pi$-twice continuously differentiable periodic parameter curve with the non-vanishing first derivative $\eta'_j(t) \neq 0$.

As shown in Figure A1, if $p_j \neq 0$, $\Gamma_j$ can be split into $p_j$ segments $\Gamma_{j,i}\ i = 1,2, \dots, p_j$ by corner points $CP_{j,1}, CP_{j,2}, \dots, CP_{j,p_j}$. Segments $\Gamma_{j,i}\ i = 1,2, \dots, p_j$ could be straight lines, arcs, or splines. If $\Gamma_{j,i}$ is a straight segment or arc, $\Gamma_{j,i}$ can be parameterized with equal arc lengths as:

$$|\eta'_{j,i}(t)| = 1 \qquad t \in [0, Len] \tag{A-7}$$

where, $Len$ represents the total length of the line segment or arc. As shown in Eq. A-1, $\Gamma_{j,i}$ can be an open B-spline segment with $k$ control points and parameter $t \in [0, k-3]$. Now, the segmented smooth closed curves are parameterized in the following form:

$$\begin{cases} \eta_{j,1}(t) & t \in J_{j,1} \\ \eta_{j,2}(t) & t \in J_{j,2} \\ \vdots \\ \eta_{j,p_j}(t) & t \in J_{j,p_j} \end{cases} \tag{A-8}$$

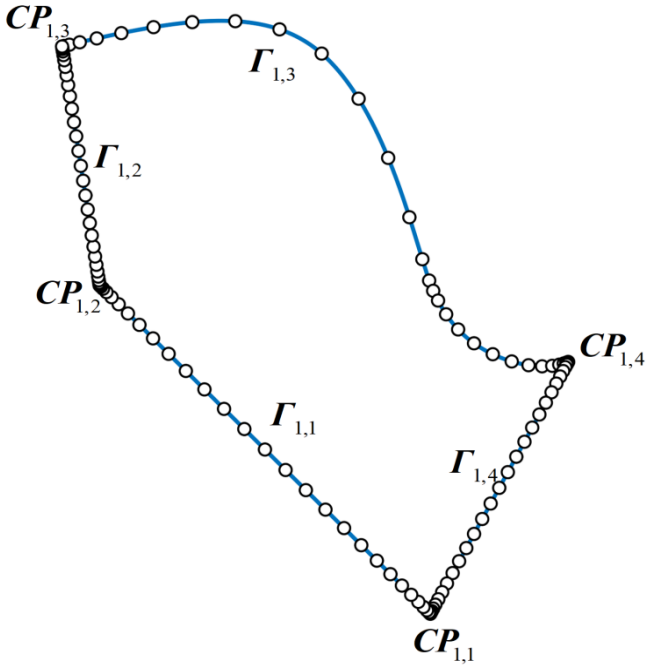

**Figure A1.** The boundary $\Gamma_1$ in Fig. 2(a) are discretely into 128 points.

Then, we use $\hat{t}$ instead of $t$ as the segment curve parameter as the form $\eta_{j,i}(t(\hat{t}))$ and $t(\hat{t})$ can be represented by:

$$t(\hat{t}) = \left(\hat{t} - \frac{2\pi i}{p_j}\right) sup(J_{j,i}) \qquad \hat{t} \in \left[\frac{2\pi i}{p_j}, \frac{2\pi(i+1)}{p_j}\right] \tag{A-9}$$

where $sup(\cdot)$ represents the upper bound of the interval. Now, the segmented smooth closed curves are parameterized as:

$$\begin{cases} \eta_{j,0}(\hat{t}) & \hat{t} \in \left[0, \frac{2\pi}{p_j}\right] \\ \eta_{j,1}(\hat{t}) & \hat{t} \in \left[\frac{2\pi}{p_j}, \frac{4\pi}{p_j}\right] \\ \vdots \\ \eta_{j,p_j-1}(\hat{t}) & \hat{t} \in \left[\frac{2\pi(p_j-1)}{p_j}, 2\pi\right] \end{cases} \quad \text{(A-10)}$$

Hence, whole $\eta_j(\hat{t})$  $\hat{t} \in [0, 2\pi]$ is parameterized with $2\pi$-periodic parameters. But grading mesh parameter transformations are still needed to make arc length changes more slowly with $t$ around corners, which allows the later solution of the slit map to converge.

Suppose that $\sigma(t)$ represents the bijective, strictly monotonically increasing and infinitely differentiable function maps with $\sigma(t): [0, 2\pi] \to [0, 2\pi]$, $\sigma'(0) = \sigma'(2\pi) = 0$ defined by(Nasser et al., 2008; Kress, 1990):

$$\sigma(t) = 2\pi \frac{[v(t)]^p}{[v(t)]^p + [v(2\pi-t)]^p} \quad \text{(A-11)}$$

where $p \geq 2$ is the constant grading parameter, and $v(t)$ is in the following form:

$$v(t) = \left(\frac{1}{p} - \frac{1}{2}\right)\left(\frac{\pi-t}{\pi}\right)^3 + \frac{1}{p}\frac{t-\pi}{\pi} + \frac{1}{2} \quad t \in [0, 2\pi] \quad \text{(A-12)}$$

$\eta_j(\hat{t})$ can be re-parameterized by $\hat{\hat{t}}$ as(Liesen et al., 2017):

$$\hat{\hat{t}} = \begin{cases} \frac{\sigma(p_j \hat{t})}{p_j} & \hat{t} \in \left[0, \frac{2\pi}{p_j}\right] \\ \frac{\sigma\left(p_j\left(\hat{t} - \frac{2\pi}{p_j}\right)\right)}{p_j} + \frac{2\pi}{p_j} & \hat{t} \in \left[\frac{2\pi}{p_j}, \frac{4\pi}{p_j}\right] \\ \vdots \\ \frac{\sigma\left(p_j\left(\hat{t} - \frac{2\pi(p_j-1)}{p_j}\right)\right)}{p_j} + \frac{2\pi(p_j-1)}{p_j} & \hat{t} \in \left[\frac{2\pi(p_j-1)}{p_j}, 2\pi\right] \end{cases} \quad \text{(A-13)}$$

After converting the algebraic symbol $\hat{\hat{t}}$ to $t$, we have $\eta_j(t)$  $t \in J_j = [0, 2\pi]$ with $\eta_j\left(i\frac{2\pi}{p_j}\right) = CP_{j,i}$ and $\eta_j'\left(i\frac{2\pi}{p_j}\right) = 0$, where $i = 0, 1, \ldots, p_j$. As shown in Figure A1, the white points are the points where the parameters are evenly divided, and the discrete points near the corners are denser. Since then, $2\pi$-periodic parameterization can be performed on all the smooth or segmentally smooth boundaries $\Gamma$ by $\eta$:

$$\eta(t,j) = \eta_j(t) \quad t \in J_j = [0, 2\pi] \quad j = 0,1,\ldots,m \tag{A-14}$$

A-2. Solving slit mapping on boundaries

The total parameter domain $J$ of $\eta$ is the disjoint union of $J_0, J_1, \ldots J_m$:

$$J = \bigsqcup_{j=0}^{m} J_j = \bigcup_{j=0}^{m} \{(t,j) \ t: \in J_j\}$$

Thus, $J$ is the domain with the space $[0, 2\pi] \times \{0,1,\ldots,m\}$, and we default the $j$ corresponds to $t$ in binary parameter $(t,j)$. Thus, $\eta$ is in the following disjoint parameter form:

$$\eta(t) = \begin{cases} \eta_0(t) & t \in J_0 = [0, 2\pi] \\ \vdots \\ \eta_m(t) & t \in J_m = [0, 2\pi] \end{cases} \tag{A-15}$$

We define the complex-valued function $A_j$ on $\Gamma_j$ by:

$$A_j(\eta_j) = \begin{cases} e^{i\left(\frac{\pi}{2} - \theta_j(\eta_j)\right)} \eta_j & \text{If } G \text{ is bounded} \\ e^{i\left(\frac{\pi}{2} - \theta_j(\eta_j)\right)} & \text{If } G \text{ is unbounded} \end{cases} \quad j = 0,1\ldots m \tag{A-16}$$

where $\theta_j$ presents oblique angles defined on $\Gamma_j$. Since we only use two typical types of slit maps with bounded $G$ and constant oblique angles (see Eq. 1.3 in (Nasser, 2015)), $\theta_j = \frac{\pi}{2}$, $j = 0,1,\ldots,m$, so only the complex-valued function $A_j$ needs to be defined on $\Gamma_j$ by $A_j = \eta_j$, which means $A(t) = \eta(t) \ t \in J$.

Then, the generalized Neumann kernel $N(s,t)$ is defined on space $J \times J$ by:

$$N(s,t) = \frac{1}{\pi} Im\left(\frac{A_s \eta'_t}{A_t(\eta_t - \eta_s)}\right) = \frac{1}{\pi} Im\left(\frac{\eta_s \eta'_t}{\eta_t(\eta_t - \eta_s)}\right) \quad (s,t) \in J \times J \tag{A-17}$$

The kernel $N(s,t)$ is continuous with:

$$N(t,t) = \frac{1}{\pi}\left(\frac{1}{2} Im\left(\frac{\eta''_t}{\eta'_t}\right) - Im\left(\frac{A'_t}{A_t}\right)\right) = \frac{1}{\pi}\left(\frac{1}{2} Im\left(\frac{\eta''_t}{\eta'_t}\right) - Im\left(\frac{\eta'_t}{\eta_t}\right)\right) \tag{A-18}$$

Also, a real kernel $M(s,t)$ is defined on space $J \times J$ by:

$$M(s,t) = \frac{1}{\pi} Re\left(\frac{A_s \eta_t'}{A_t(\eta_t - \eta_s)}\right) = \frac{1}{\pi} Re\left(\frac{\eta_s \eta_t'}{\eta_t(\eta_t - \eta_s)}\right) \quad (s,t) \in J \times J \tag{A-19}$$

In each subspace $J_j \times J_j$ $j = 1,2,\ldots,m$:

$$M(s,t) = -\frac{1}{2\pi} \cot\left(\frac{s-t}{2}\right) + M_1(s,t) \quad (s,t) \in J_j \times J_j \quad j = 1,2,\ldots,m \tag{A-20}$$

where $M_1$ is a continuous kernel with following form:

$$M_1(t,t) = \frac{1}{\pi}\left(\frac{1}{2} Re\left(\frac{\eta_t''}{\eta_t'}\right) - Re\left(\frac{A_t'}{A_t}\right)\right) = \frac{1}{\pi}\left(\frac{1}{2} Re\left(\frac{\eta_t''}{\eta_t'}\right) - Re\left(\frac{\eta_t'}{\eta_t}\right)\right) \tag{A-21}$$

The Fredholm integral operator $N$ with a continuous kernel $N(s,t)$ can be defined by:

$$N\mu(s) = \int_J M(s,t)\mu(t) dt \quad t \in J, (s,t) \in J \times J \tag{A-22}$$

The singular integral operator $M$ with a cotangent singularity-type kernel $M(s,t)$ can be defined by:

$$M\mu(s) = \int_J M(s,t)\mu(t) dt \quad t \in J, (s,t) \in J \times J \tag{A-23}$$

Now, solving the two types of slit mapping $\omega(z)$ analytic in $G$ shown in Figure 2 is transformed into the Riemann-Hilbert problem in the following form:

$$Re[Af] = \gamma \text{ on } \Gamma \tag{A-24}$$

where $f$ is the intermediate function for solving $\omega(z)$ on $\Gamma$, and $\gamma$ is a real-valued function defined on $\Gamma$. For disc slit mapping, where $\omega: \bar{G} \to \bar{D}$, we define real-valued $\gamma$ on $\Gamma$ by:

$$\gamma(t) = -Re[log(\eta(t))] \tag{A-25}$$

For annular slit mapping, where $\omega: \bar{G} \to \bar{A}$, we define real-valued $\gamma$ on $\Gamma$ by:

$$\gamma(t) = -Re\left[log\left(1 - \frac{\eta(t)}{z_1}\right)\right] \tag{A-26}$$

Eq. A-24 is not uniquely solvable for the winding number $\hbar = 1$ in our cases(see Eq. 1.2 in (Nasser, 2015)), but a unique undetermined real piecewise function $h(t) = C_j \, t \in J_j \, j = 1,2,\ldots,m, C_j \in \mathbb{R}$ can be found, which makes the following Riemann-Hilbert problem is uniquely solvable:

$$Re[Af] = \gamma + h \text{ on } \Gamma \tag{A-27}$$

Let

$$\mu = Im[Af] \text{ on } \Gamma \tag{A-28}$$

Thus, $f$ can be calculated by:

$$f = \frac{\gamma + h + i\mu}{A} \text{ on } \Gamma \tag{A-29}$$

The function $\mu$ is the unique solution to the integral equation:

$$\mu(t) - N(s,t)\mu(t) = -M(s,t)\gamma(t) \quad t \in J, (s,t) \in J \times J \tag{A-30}$$

Firstly, the boundary curve is parameterized by Eq. A-14, and then, $J$ is discretized by evenly divided parameters on each subinterval $J_j$ $j = 0,1,2 \ldots m$ into $n$ points:

$$J = \bigcup_{j=0}^{m} \bigcup_{k=0}^{n-1} \left(\frac{2\pi k}{n}, j\right)$$

Thus, $J$ and $J \times J$ are discretized into $(m+1)n$ and $(m+1)^2 n^2$ points, respectively. We denote each discretized element $t_p \in J$ or $s_p \in J$ with $p = jn + k$. By using the trapezoidal rule and Wittich's discretize method, Eq. A-30 can be written in the form of a matrix equation:

$$\begin{bmatrix} \mu(t_1) \\ \mu(t_2) \\ \vdots \\ \mu(t_{(m+1)n}) \end{bmatrix} - \frac{2\pi}{n} \begin{bmatrix} \widehat{N}(s_1,t_1) & \widehat{N}(s_1,t_2) & \cdots & \widehat{N}(t_1,t_{(m+1)n}) \\ \widehat{N}(s_2,t_1) & \ddots & & \widehat{N}(s_2,t_{(m+1)n}) \\ \vdots & & \ddots & \vdots \\ \widehat{N}(s_{(m+1)n},t_1) & & \cdots & \widehat{N}(s_{(m+1)n},t_{(m+1)n}) \end{bmatrix} \begin{bmatrix} \mu(t_1) \\ \mu(t_2) \\ \vdots \\ \mu(t_{(m+1)n}) \end{bmatrix} =$$

$$-\frac{2\pi}{n} \begin{bmatrix} \widehat{M}(s_1,t_1) & \widehat{M}(s_2,t_1) & \cdots & \widehat{M}(t_1,t_{(m+1)n}) \\ \widehat{M}(s_2,t_1) & \ddots & & \widehat{M}(s_j,t_{(m+1)n}) \\ \vdots & & \ddots & \vdots \\ \widehat{M}(s_{(m+1)n},t_1) & & \cdots & \widehat{M}(s_{(m+1)n},t_{(m+1)n}) \end{bmatrix} \begin{bmatrix} \gamma(t_1) \\ \gamma(t_2) \\ \vdots \\ \gamma(t_{(m+1)n}) \end{bmatrix} \tag{A-31}$$

in which:

$$\widehat{N}(s_i, t_j) = \begin{cases} 0 & i = j \\ N(s_i, t_j) & i \neq j \end{cases} \tag{A-32}$$

$$\widehat{M}(s_i, t_j)
= \begin{cases} 0 & \text{for } i = j \\ M_1(s_i, t_j) & \text{for } i/n = j/n \text{ and } (rem(i,n) - rem(j,n)) \text{ is even} \\ M_1(s_i, t_j) - \frac{1}{\pi}\cot\left(\frac{s-t}{2}\right) & \text{for } i/n = j/n \text{ and } (rem(i,n) - rem(j,n)) \text{ is odd} \\ M(s_i, t_j) & \text{otherwise} \end{cases}$$

(A-33)

where / represents the rounding-down operator, and $rem(i, n)$ represents the remainder of $i$ divided by $n$. Nasser has proposed a method for fast solving $(m + 1)n$-ordered linear equations in Eq. A-31, and see (Nasser, 2015) for details.

After the value of $\mu$ is obtained, $h$ can be given by:

$$h = [M\mu - (I - N)\gamma]/2 \tag{A-34}$$

After the values of $\mu$ and $h$ are obtained, $f$ can be given by Eq. A-29.

For the disc slit mapping, we want $\omega(z)$ that maps $\Gamma_0$ onto the unit circle $R_0 = 1$ and maps $\Gamma_1, \Gamma_2, \ldots, \Gamma_m$ onto circular arc slit with pending radii $R_1, R_2, \ldots, R_m$. Then, the boundary values of the mapping function $\omega$ satisfy:

$$Re\left[\ln\left(\omega(\eta_j(t))\right)\right] = \begin{cases} 0, & j = 0 \\ \ln(R_j), & 0 < R_j < 1, \quad j = 1,2,\ldots,m \end{cases} \tag{A-35}$$

If $\omega(0) = 0$ and $\omega'(0) > 0$ are set as restrictions, $\omega(z)$ has a unique solution in the following form:

$$\omega(z) = cze^{zf(z)} \qquad z \in \Gamma \tag{A-36}$$

For the annular slit mapping, we want $\omega(z)$ that maps $\Gamma_0$ onto the unit circle with $R_0 = 1$, maps $\Gamma_1$ onto the circle with $R_1 < 1$, and maps $\Gamma_2, \Gamma_3, \ldots, \Gamma_m$ onto circular arc slit with pending radii $R_2, R_3, \ldots R_m$. Then, the boundary values of the mapping function $\omega$ also satisfy Eq. A-35.

If $\omega(0) > 0$ is set as a restriction, $\omega(z)$ has a unique solution in the following form:

$$\omega(z) = c\left(1 - \frac{z}{z_1}\right)e^{zf(z)} \qquad z \in \Gamma \tag{A-37}$$

where $c = e^{h(\eta_0)}$ in Eq. A-36 and Eq. A-37.

A-3. Solving slit mapping and inverse slit mapping not on boundaries

The internal points $z \in G$, $\omega(z)$ can be solved by the Cauchy integral formula:

$$\omega(z) = \frac{\sum_{j=0}^{m}\int_0^{2\pi}\frac{\omega(\eta_j(t))\eta_j'(t)}{\eta_j(t)-z}dt}{\sum_{j=0}^{m}\int_0^{2\pi}\frac{\eta_j'(t)}{\eta_j(t)-z}dt} \qquad z \in G \tag{A-38}$$

For internal mapping points $w \in D$ or $w \in A$, the inverse map $z = \omega^{-1}(w)$ can also be calculated by the Cauchy integral formula:

$$\omega^{-1}(w) = \frac{1}{2\pi i}\sum_{j=0}^{m}\int_0^{2\pi}\frac{\eta_j(t)\eta_j'(t)\omega_j'(\eta_j)}{\omega_j(t)-w}dt \qquad w \in D \text{ or } w \in A \tag{A-39}$$

The discrete form of the integral calculation in Eq. A-38 and Eq. A-39 can be accelerated by the fast multipole method, see (Nasser, 2015) for details.